\documentclass{llncs}
\usepackage{times} \usepackage{graphicx}
\usepackage{latexsym} \usepackage{amsmath, amssymb}
\usepackage{url}
\begin{document}
\title{Stochastic Constraint Programming\\
as Reinforcement Learning}
\author{S. D. Prestwich${}^1$, R. Rossi${}^2$, S. A. Tarim${}^3$\\
${}^1$Insight Centre for Data Analytics,
University College Cork, Ireland\\
${}^2$University of Edinburgh Business School, Edinburgh, UK\\
${}^3$Department of Management, Cankaya University, Ankara, Turkey
}
\institute{\tt s.prestwich@cs.ucc.ie, roberto.rossi@ed.ac.uk, at@cankaya.edu.tr}
\maketitle
\begin{abstract}

Stochastic Constraint Programming (SCP) is an extension of Constraint
Programming (CP) used for modelling and solving problems involving
constraints and uncertainty.  SCP inherits excellent modelling
abilities and filtering algorithms from CP, but so far it has not been
applied to large problems.  Reinforcement Learning (RL) extends
Dynamic Programming to large stochastic problems, but is
problem-specific and has no generic solvers.  We propose a hybrid
combining the scalability of RL with the modelling and constraint
filtering methods of CP.  We implement a prototype in a CP system and
demonstrate its usefulness on SCP problems.

\end{abstract}

\section{Introduction}

Stochastic Constraint Programming (SCP) is an extension of Constraint
Programming (CP) designed to model and solve complex problems
involving uncertainty and probability, a direction of research first
proposed in \cite{BenEtc,Wal}.  SCP problems are in a higher
complexity class than CP problems and can be much harder to solve, but
many real-world problems contain elements of uncertainty so this is an
important class of problems.  They are traditionally tackled by
Stochastic Programming \cite{BirLou}, but a motivation for SCP is that
it should be able to exploit the richer choice of variables and
constraints used in CP, leading to more compact models and the use of
powerful filtering algorithms.

However, so far SCP has not been applied to very large problems.  If a
problem has many decision variables we can apply metaheuristics
\cite{PreEtc2,PreEtc3,PreEtc5,Wal} but we must still check all
scenarios to obtain an exact solution, though in special cases a
subset is sufficient \cite{PreEtc}.  If we are content with an
approximate solution we can apply scenario reduction by sampling
\cite{DupEtc} or approximation \cite{BiaEtc}, but scenario reduction
methods can be nontrivial to analyse and apply.  Confidence intervals
can be applied to control approximations \cite{RosEtc2} but this does
not address the issue of scaling up to a huge number of scenarios.  In
summary, to solve large real-world problems via SCP one must think
carefully about scenario reduction, and to do so can require
significant mathematical expertise.  Moreover, the number of scenarios
required might turn out to be unmanageable.

In contrast, many large stochastic and adversarial problems have been
successfully solved by methods from Reinforcement Learning (RL)
\cite{SutBar}, which is related to Neuro-Dynamic Programming
\cite{BerTsi} and Approximate Dynamic Programming \cite{Pow}.  RL
algorithms are designed for problems in which rewards may be delayed,
so that the consequences of making a decision are not known until a
later time.  RL algorithms such as SARSA and Q-Learning can be used to
find high-quality solutions to large-scale problems.  In RL
researchers are less concerned with sample sizes, confidence intervals
or other statistical issues.  Typically they model their problem,
choose an RL algorithm and tune it to their application.  These
methods have been successfully applied to problems in robotics,
control, game playing, trading and human-computer interfaces, for
example.  Perhaps most famously, RL was used to learn how to play the
game of Backgammon \cite{Tes} by trial-and-error self-play and without
human intervention, leading to a world-class player.  Related methods
developed in Operations Research to handle exponentially-many actions
are able to handle far larger problems, for example the scheduling of
tens of thousands of trucks \cite{Pow}.  Such applications show that
the solutions found by RL can be good enough for practical purposes.

Such applications are far beyond the scope of current SCP techniques.
Our aim is to boost the scalability of SCP so that it can tackle
similar problems to RL, while retaining its modelling power and
constraint filtering techniques.  From the RL point of view, this is
of interest because it provides a generic RL solver for a significant
class of problems, which uses constraint filtering to reduce the size
of the state space.  Section \ref{CP} provides background on SCP,
Section \ref{method} describes our method, Section \ref{experiments}
presents experimental results using an implementation in a CP system,
and Section \ref{conclusion} draws conclusions and discusses future
work.

\section{Stochastic Constraint Programming} \label{CP}

An $m$-stage SCSP is defined as a tuple $(V,S,D,P,C,\theta,L)$ where
$V$ is a set of decision variables, $S$ a set of stochastic variables,
$D$ a function mapping each element of $V \cup S$ to a domain of
values, $P$ a function mapping each variable in $S$ to a probability
distribution, $C$ a set of constraints on $V \cup S$, $\theta$ a
function mapping each constraint in $C$ to a threshold value $\theta
\in (0,1]$, and $L=[\langle V_1,S_1 \rangle, \ldots, \langle V_m,S_m
    \rangle]$ a list of {\it decision stages\/} such that the $V_i$
  partition $V$ and the $S_i$ partition $S$.  Each constraint must
  contain at least one $V$ variable, a constraint with threshold
  $\theta(h)=1$ is a {\it hard constraint\/}, and one with
  $\theta(h)<1$ is a {\it chance constraint\/}.  To solve an $m$-stage
  SCSP an assignment to the variables in $V_1$ must be found such
  that, given random values for $S_1$, assignments can be found for
  $V_2$ such that, given random values for $S_2,\ldots$ assignments
  can be found for $V_m$ such that, given random values for $S_m$, the
  hard constraints are each satisfied and the chance constraints
  (containing both decision and stochastic variables) are satisfied in
  the specified fraction of all possible {\it scenarios\/} (set of
  values for the stochastic variables).

An SCSP solution is a {\it policy tree\/} of decisions, in which each
node represents a value chosen for a decision variable, and each arc
from a node represents the value assigned to a stochastic variable.
Each path in the tree represents a different possible scenario and the
values assigned to decision variables in that scenario.  A {\it
satisfying policy tree\/} is a policy tree in which each chance
constraint is satisfied with respect to the tree.  A chance constraint
$h \in C$ is satisfied with respect to a policy tree if it is
satisfied under some fraction $\phi \ge \theta(h)$ of all possible
paths in the tree.

\section{SCP as RL} \label{method}

In this section we describe our hybrid approach to solving SCP
problems.

\subsection{Reinforcement Learning}

RL is an area of machine learning with roots in dynamic programming,
Monte Carlo methods, optimal control and behavioural psychology.  It
is one of the three main classes of machine learning, the other two
being supervised and unsupervised learning.  RL involves the
interaction between a decision-making {\it agent\/} and its {\it
  environment\/}.  The agent seeks to optimise an expected total {\it
  reward\/} under uncertainty about its environment.  The agent can
take {\it actions\/} which may affect the future state of the
environment, which in turn may affect the agent's later options.

Rewards might be random, which is why the agent maximises their
expectation.  Rewards may also be delayed in time, so that choosing
actions involves taking into account their later consequences.  For
example when playing a game the only reward might occur at the end of
the game: 1 for a win and 0 for a loss.  Thus the agent must learn how
to react to any possible game state in order to maximise its
probability of a win.

Any state might have an associated reward.  The agent must learn a
{\it policy\/} (a function from states to actions) that maximises the
total expected reward, under the assumption that it follows an optimal
path.  To do this it estimates the total expected reward starting from
each state, typically storing the estimates in a table of {\it state
  values\/} (or in some algorithms {\it state-action values\/}).  The
agent learns these estimates by performing a large number of Monte
Carlo-style simulations called {\it episodes\/}, and updating the
values at each state encountered.

\subsection{Modelling} \label{model}

We model an SCP problem as an RL problem as shown in Table
\ref{corres}.  In this approach we can benefit from constraint
filtering methods: stronger filtering restricts our choice of actions
(domain values), so we avoid visiting more states, which may enable a
simpler state aggregation method to emulate a more complex one.  But
it is possible to reach a dead-end state in which no actions remain,
because of SCP {\it domain wipe-out\/}.  We need RL to learn to take
decisions that will avoid dead-ends, so we reward each (decision or
random) variable assignment with a constant $K$, which will typically
be greater than the greatest possible objective value.  Instead we
could relax sufficient constraints to prevent dead-ends then penalise
any violations, but this loses the advantage of constraint filtering.

\begin{table}
\begin{center}
\begin{tabular}{|l|l|}
\hline
SCP & RL\\
\hline
assigning a decision variable & action\\
assigning a random variable & environmental response\\
& moving to a new state\\
constraint & environmental response\\
& restricting future actions\\
empty assignment & initial state\\
partial assignment & state\\
complete assignment & terminal state\\
assigning all variables in turn & episode\\
objective & reward\\
{\it new feature\/} & state aggregation\\
\hline
\end{tabular}
\end{center}
\label{corres}
\caption{SCP modelled as RL}
\end{table}

From the RL point of view, the CP solver is now part of the policy.
For example, suppose we have an SCP problem containing an {\tt
  alldifferent} global constraint such as that in \cite{Reg}, and we
solve it, obtaining a policy.  If we then try to use that policy to
choose a sequence of actions, but with a different CP solver that
implements {\tt alldifferent} as a set of pairwise disequality
constraints, the different level of filtering leads to a different
state space, we will follow a different policy, and the results will
be unpredictable.  We must therefore use exactly the same CP solver
when finding a policy and using it.

This framework can handle a single chance constraint, plus any number
of hard constraints, by attempting to maximise the probability that
the constraint is satisfied: if this is greater than the threshold
then we have a solution.  However, it does not handle multiple chance
constraints.  It might be possible to extend it to chance constraints
but we do not see this as vital.  Most SP problems do not use chance
constraints, instead penalising constraint violations and minimising
the total penalty as part of the objective.  For a recent discussion
on penalty functions versus chance constraints see \cite{Bra}.

A potential problem with this scheme is that domain values for a
random variable might be filtered because of earlier decision variable
assignments, or assigning a value to a random variable might fail
because of domain wipe-out.  This would artificially rule out some
scenarios and make the solver incorrect, but it can be avoided by a
cheap runtime check: on encountering a random variable, check that it
still has its original domain; and after selecting a domain value,
check that the assignment is successful.  If the check fails, the
probe halts at the random variable.



\subsection{Solving}

The RL algorithm we shall use is a form of tabular TD(0) \cite{SutBar}
with a reward computed at the end of each episode.  However, many
problems have far too many possible states to use RL in tabular form.
To extend RL to cope with such problems researchers have applied {\it
  function approximation\/}, also referred to as {\it state
  generalisation\/} or {\it state aggregation\/}.  This is key to the
success of RL on real-world problems and we shall use it below.  To
apply our algorithm a user must provide an SCP model, including a
real-valued function on total assignments defining a reward.

\section{Experiments} \label{experiments}

We now perform experiments to evaluate our approach, which we refer to
as TDCP, on stochastic problems.  It is implemented in the Eclipse
constraint logic programming system \cite{AptWal} and all experiments
are performed on a 2.8 GHz Pentium 4 with 512 MB RAM.

\subsection{An artificial single-stage problem} \label{artificial}

As a first experiment we design an artificial single-stage problem
with known optimum solution.  The problem has $N$ decision variables
$d_i$ and $N$ random variables $r_i$ all with domain $\{1, \ldots,
N\}$.  We post an {\tt alldifferent} constraint on the decision
variables: there are $N$ variables with $N$ values, so the solution
must be a permutation of $\{1,\ldots,N\}$.  All random variable
domains have the same uniform probability distribution: each value has
probability $1/N$.  The objective is to maximise the sum of the
probabilities that each decision variable $d_i$ is no greater than
each random variable $r_{i+1} \ldots r_N$: see Figure \ref{scop},
where $\mbox{reify}(c)$ is 1 if condition $c$ is true and 0 if it is
false.  The sum of the reified terms (without expectation) is the TDL
reward.  The optimal solution is known to be $\{d_1=1, d_2=2 \ldots,
d_N=N\}$ with objective value $N(2N-1)/6$.  There are $N^N$ scenarios
so this problem cannot be solved by SCP methods without some form of
scenario reduction.

\begin{figure}[t!]
\begin{center}
\framebox{
\begin{tabular}{l}
\mbox{{\bf Maximise:}}\\
$\;\;\;\sum_{i=1}^N \sum_{j=i+1}^N \mathbb{E}\left[\mbox{reify}(d_i \le r_j)\right]$\\
\mbox{{\bf Constraints:}}\\
$\;\;\;\mbox{\tt alldifferent}(\{d_1, \ldots, d_N\})$\\
\mbox{{\bf Decision variables:}} \\
$\;\;\;d_1 \ldots d_N \in \{1,2, \ldots, N\}$\\
\mbox{{\bf Random variables:}} \\
$\;\;\;r_1 \ldots r_N \in \{1(1/N), \ldots, N(1/N)\}$\\
\mbox{{\bf Stage structure:}} \\
\begin{tabular}{l@{\hspace{3mm}}l}
$\;\;\;V_1=\{d_1, \ldots, d_N\}$ & $S_1=\{r_1, \ldots, r_N\}$\\
\multicolumn{2}{l}{$\;\;\;L=[\langle V_1,S_1\rangle]$}
\end{tabular}
\end{tabular}
}
\end{center}
\caption{An artificial single-stage SCP problem}
\label{scop}
\end{figure}

To handle the exponentially large number of states we use a form of
state aggregation based on {\it Zobrist hashing\/} \cite{Zob} with $H$
hash table entries for some integer $H$, which works as follows.  To
each (decision or random) variable-value pair $\langle v,i \rangle$ we
assign a random integer $r_{vi}$ which remains fixed.  At any point
during an episode we have some set $S$ of assignments $\langle v,i
\rangle$, and we take the exclusive-or of the $r_{vi}$ values
associated with these assignments:
\[
X_S = \bigoplus_{\langle v,i \rangle \in S} r_{vi}
\]
Finally, we use $X_S \!\!\!\mod\! H$ as an index to an array $V$ with
$H$ entries.  So at any point during an episode we are in an RL state
with variables $S$ assigned, and we use array element $V[X_S
  \!\!\!\mod\!  H]$ as the state value estimate.  If $H$ is
sufficiently large then hash collisions are unlikely, and we will have
a unique array element for each state encountered.  In practice some
hash collisions will occur, leading to multiple states sharing value
estimates, and less exact results.  Nevertheless, we shall show
empirically that good results can be obtained.  Our hash-based state
aggregation can also be applied to other single-stage SCP problems, or
multi-stage problems in which recourse actions are computed by an
algorithm other than RL (as in the problem of Section \ref{quake}).
However, we do not expect it to be successful on all multi-stage
problems.

\begin{figure}[t!]
\begin{center}
\includegraphics[scale=0.68]{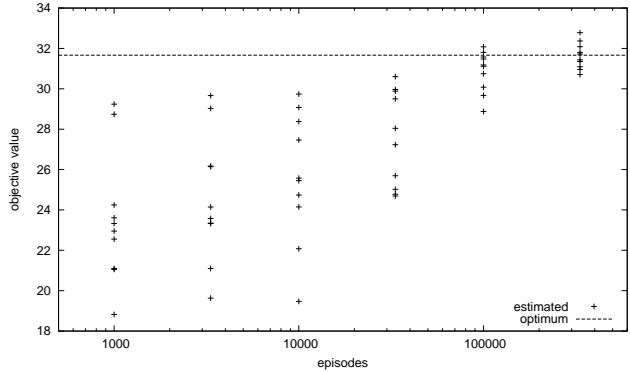}
\end{center}
\caption{Results for TD(0) on the SCP problem}
\label{results}
\end{figure}

The scatter plot in Figure \ref{results} shows the results for $N=10$
using $H=10^5$.  For several numbers of episodes (all far less than
the full ten billion) we run TD(0) ten times with different random
seeds.  The graph shows that as more episodes are used for learning,
the estimated objective function value converges to the known optimum
value.

\subsection{Pre-disaster planning} \label{quake}

In this section we tackle a pre-disaster planning problem introduced
by Peeta {\it et al.\/} \cite{PeeEtc} who solved it approximately
using a Monte Carlo method with function approximation.  The six
problem instances were later solved exactly in \cite{PreEtc}.  A
detailed description of the problem can be found in those papers, and
here we state only its main features.

This is a two-stage problem in which the recourse action is determined
by solving a shortest path problem.  The first stage has 30 binary
decision variables representing investment in links of a
transportation network, and 30 binary random variables representing
the survival or failure of those links in a hypothetical earthquake,
according to given survival probabilities.  The probabilities are
assumed to be independent of each other, but they depend on the
investment decisions: decision-dependent probabilities are a
non-standard feature of SP and SCP called {\it endogenous
  uncertainty\/}.  (This can make problems harder to solve by some
methods, but not for a simulation-based approach such as ours.)  We
can choose to invest in any subset of the links subject to a budget
constraint, and three alternative budget levels $B_1<B_2<B_3$ are
chosen.  The objective is to minimise the expected total path length
between five pairs of nodes in the network, where a penalty $M$ is
imposed when no path exists between a pair of nodes.  Two alternative
penalty schemes are used, which we shall refer to as low-{\it M\/} and
high-{\it M\/}, giving a total of six problem instances.

An SCP model is shown in Figure \ref{networkeg}.  For each link $e \in
E$ (where $E$ is the set of links in the network) we define a binary
decision variable $y_e$ which is 1 if we invest in that link and 0
otherwise.  We define a binary stochastic variable $r_e$ which is 1 if
link $e$ survives and 0 if it fails.  We define a single second-stage
decision variable $z$ to be computed by a shortest-path algorithm.
Following Peeta {\it et al.\/} we denote the survival (non-failure)
probability of link $e$ by $p_e$ without investment and $q_e$ with,
the investment required for link $e$ by $c_e$, the length of link $e$
by $t_e$, the budget by $B$, and the penalty for no path from source
to sink by $M$.  $\mbox{\tt shortest\_path\_cost}(M,\{t_e|e \in
E\},\{r_e|e \in E\},z)$ is a global constraint that constructs a
representation of the graph from the $r_e$ values, uses Dijkstra's
algorithm to find a shortest path between source and sink, and
computes its length $z$; if source and sink are unconnected then
$z=M$.  We implemented this constraint via an Eclipse {\it suspended
  goal\/} whose execution is delayed until the second stage.  To model
failure probabilities we define real auxiliary decision variables
$f_e$.  The $f_e$ are constrained to be $1-p_e$ if link $e$ is
invested in ($y_e=1$) and $1-q_e$ otherwise.  Because they are
auxiliary variables and functionally dependent on the $y_e$ we do not
include them in the stage structure.

\begin{figure}[t!]
\begin{center}
\framebox{
\begin{tabular}{ll}
\mbox{{\bf Minimise:}}\\
$\;\;\;\mathbb{E}\{z\}$\\
\mbox{{\bf Constraints:}}\\
$\;\;\;c_1:\,\sum_{e \in E} c_e y_e \le B$ &\\
$\;\;\;c_2:\,f_e=y_e (1-q_e) +(1-y_e)(1-p_e)$ & $(\forall e \in E)$\\
$\;\;\;c_3:\,\mbox{\tt shortest\_path\_cost}(M,\{t_e|e \in E\},\{r_e|e \in E\},z)$ &\\
\mbox{{\bf Decision variables:}} \\
$\;\;\;y_e \in \{0,1\},\,f_e \in \mathbb{R}$ & $(\forall e \in E)$\\
$\;\;\;z \in \mathbb{R}$ &\\
\mbox{{\bf Random variables:}} \\
$\;\;\;r_e \in \{0(f_e),1(1-f_e)\}$ & $(\forall e \in E)$\\
\mbox{{\bf Stage structure:}} \\
\begin{tabular}{l@{\hspace{3mm}}l}
$\;\;\;V_1=\{y_e\,|\,e \in E\}$ & $S_1=\{r_e\,|\,e \in E\}$\\
$\;\;\;V_2=\{z\}$ & $S_2=\emptyset$\\
\multicolumn{2}{l}{$\;\;\;L=[\langle V_1,S_1\rangle,\langle V_2,S_2\rangle]$}
\end{tabular}
\end{tabular}
}
\end{center}
\caption{An SCP model for the pre-disaster planning problem.}
\label{networkeg}
\end{figure}

The problem is hard to solve exactly by standard SP and SCP methods,
partly because of its endogenous uncertainty, but mainly because it
has approximately a billion ($2^{30}$) scenarios.  Peeta {\it et
  al.\/} therefore used function approximation and Monte Carlo
simulation to find good solutions.  However, a symmetry-based
technique called {\it scenario bundling\/} was later applied to find
exact solutions \cite{PreEtc}.

This is an ideal test problem for our approach: it is an interesting
stochastic optimisation problem based on real-world data; it is a
large, hard problem (unless we use scenario bundling); unusually for
such a problem we know the exact answer (via scenario bundling); again
unusually we can exactly evaluate new solutions (via scenario
bundling); and we can compare our approximate results with those found
by another approximate approach (that of Peeta {\it et al.\/}).  We
again use our Zobrist hashing technique from Section \ref{artificial}.
Though this problem is two-stage because it has recourse actions, in a
sense it is only a one-stage problem because the recourse actions are
computed by a shortest path algorithm: RL need not learn how to react
to different scenarios. This makes the problem appropriate for our
hashing technique.  In a true multi-stage problem TDCP will never
learn how to react to any given scenario because it is unlikely ever
to encounter the same scenario twice.

In experiments we found quite different solution quality in different
runs, so for each problem instance we performed ten runs of TDCP and
report the best results in Table \ref{quake1results}.  We show the
optimum objective values from \cite{PreEtc}, the exact evaluation of
the plans found by Peeta {\it et al.\/}, the TDCP plans (a list of
the links invested in), their TDCP-estimated objective values and
their exact values.  Each run of Peeta {\it et al.\/} took
approximately 380 seconds on a PC with $2 \times 2.8$ GHz Xeon
processor and 5 GB RAM implemented in Matlab 7.0, while ours took
approximately 1000 seconds each on a roughly comparable machine: we
are unable to compare execution times directly but ours seem
reasonably efficient.

\begin{table}
\begin{center}
\begin{tabular}{|rr|rr|lrr|}
\hline
$B$ & $M$ & optimum & \cite{PeeEtc} & TDCP plan & estimated & actual\\
\hline
1 & low & 83.080 & 86.717 & (4 6 21 22 25) & 83.521 & 83.796\\ 
2 & low & 66.188 & 70.035 & (6 7 12 17 20 21 22 25) & 71.968 & 72.329\\
3 & low & 57.680 & 59.532 & (1 2 4 5 7 10 12 16 20 21 22 23 25) & 62.229 & 62.283\\
1 & high & 212.413 & 215.670 & (2 4 10 21 22 25) & 219.078 & 219.358\\
2 & high & 120.080 & 121.818 & (5 10 12 17 19 20 21 22 23 25) & 128.987 & 128.543\\
3 & high & 78.402 & 87.927 & (3 4 10 12 13 17 19 20 21 22 25 28) & 84.275 & 83.988\\
\hline
\end{tabular}
\end{center}
\caption{Results for stochastic earthquake problem}
\label{quake1results}
\end{table}



The TDCP objective estimates turn out to be quite accurate, with at
most 0.5\% deviance from the actual objective value.  Our approach
required multiple runs of $10^6$ episodes instead of one run, so it
appears to be less efficient than that of Peeta {\it et al.\/}, but
the results are competitive and in two cases were closer to optimal.
It is perhaps surprising that TDCP, a general-purpose SCP algorithm
with random state aggregation, gives comparable results to the more
sophisticated and problem-specific approximation of Peeta {\it et
  al.\/}

As an illustration of an advantage of using a generic CP-based solver,
we experimented further with the model.  In principle we can apply
many standard CP techniques to improve the SCP model: add implied
constraints, change the filtering algorithm for a constraint (for
example by using a global constraint), break symmetries, exploit
dominances, experiment with different variable orderings, and so on.
For this problem we found improved results by making two changes.
Firstly, we added constraints to limit the search to maximal
solutions:
\[
By_e+z+c_e > B \;\;\; (\forall e \in E)
\]
These constraints exclude non-maximal investment plans in which we do
not invest in a link despite there being enough unspent money to do
so.  Secondly we randomly permute the $y_e$ before starting the
search: we did not find a good deterministic variable ordering, but by
randomising we hope to find a better ordering (if one exists) over
multiple runs.  We obtained some improved results: for $B$=2 $M$=low
we found the plan (1 4 10 15 17 20 21 22 25 23) with estimated
objective value 67.271 and actual value 67.334, and for $B$=1 $M$=high
plan (10 17 21 22 23 25) with estimated value 211.492 and actual value
212.413 (this is the optimal plan from \cite{PreEtc}), both better
than the plans of Peeta {\it et al.\/} However, the use of a random
variable permutation caused a greater variability in plan quality.
Clearly the variable ordering has a strong effect on the search, and
more research might find a good heuristic.  But the main point of this
experiment was to show that it is very easy to experiment with
alternative SCP models and heuristics to obtain new RL algorithms for
SCP.


%
%


\section{Conclusion} \label{conclusion}

We implemented a simple RL algorithm in a CP solver, and obtained a
novel algorithm for solving SCP problems.  We showed that this RL/CP
hybrid can find high-quality solutions to hard problems.  We believe
that exploiting Machine Learning methods is a good direction for SCP
research, to make it a practical tool for real-world problems.  In
future work we shall show that our approach extends to multistage SCP
problems using different state aggregation techniques (we have
preliminary results on an inventory control problem).

This work should also be of interest from an RL perspective.  Firstly,
implementing RL algorithms in a CP solver enables the user to perform
rapid prototyping of RL methods for new problems.  For example, simply
by specifying a different filtering algorithm for a global constraint
we obtain a new RL solver.  Secondly, we now have an RL solver for an
interesting class of problem (SCP problems).  There are no
general-purpose RL solvers available because, like Dynamic
Programming, RL is a problem-specific approach.  Thirdly, allowing the
use of constraint filtering methods in RL potentially boosts its
ability to solve tightly-constrained problems.

\subsubsection*{Acknowledgments}

This publication has emanated from research supported in part by a
research grant from Science Foundation Ireland (SFI) under Grant
Number SFI/12/RC/2289.

\bibliographystyle{plain}

\end{document}